\documentclass[letterpaper, 10 pt, conference]{ieeeconf}  

\IEEEoverridecommandlockouts                              

\usepackage{graphics} 
\usepackage{epsfig} 
\usepackage{amsmath} 
\usepackage{amssymb}  
\usepackage{mathtools} 
\usepackage[nocompress]{cite}

\usepackage[normalem]{ulem}
\usepackage{picinpar}
\usepackage{url}
\usepackage{flushend}
\usepackage{colortbl}
\usepackage{soul}
\usepackage{multirow}
\usepackage{pifont}
\usepackage{color}
\usepackage{alltt}
\usepackage{hyperref}
\usepackage{siunitx}
\usepackage{breakurl}
\usepackage{epstopdf}
\usepackage{pbox}
\usepackage{subfigure}
\usepackage{threeparttable}
\usepackage{graphicx}
\usepackage{bm} 
\usepackage{adjustbox}
\usepackage{booktabs}
\usepackage[utf8]{inputenc}
\usepackage{booktabs}
\usepackage{threeparttable}
\usepackage{textcomp}
\usepackage{hyperref}
\usepackage{booktabs}
\usepackage{float}
\usepackage{indentfirst}
\usepackage{multirow}
\usepackage{multicol}
\usepackage[table,xcdraw]{xcolor}
\usepackage{colortbl}
\usepackage{makecell}
\usepackage{arydshln}
\usepackage{xcolor} 

\usepackage{amsfonts}

\newcommand{\T}{^{\mathsf{T}}}
\definecolor{bestblue}{RGB}{183,215,228}
\definecolor{typegray}{RGB}{230,230,230}

\title{\LARGE \bf FLoTS-SLAM: Optical Flow-Derived Uncertainty for Monocular
Dynamic SLAM with Two-Stage Optimization}

\author{%
Haorui Du$^{1}$, Lipu Zhou$^{\ast 1}$%
\thanks{$^{\ast}$Corresponding author.}%
\thanks{$^{1}$The authors are with the School of Instrument Science
and Opto-electronics Engineering, Beihang University,
Beijing 100191, China
(e-mail: \texttt{duhr\_buaa@buaa.edu.cn},
\texttt{zhoulipu@buaa.edu.cn}).}%
}

\begin{document}

\maketitle
\thispagestyle{empty}
\pagestyle{empty}

\setlength{\abovedisplayskip}{4pt}
\setlength{\belowdisplayskip}{4pt}
\setlength{\abovedisplayshortskip}{4pt}
\setlength{\belowdisplayshortskip}{4pt}

\begin{abstract}
Monocular simultaneous localization and mapping (SLAM) remains challenging in
dynamic environments. Thus, real-time dynamic object identification in
such environments is a critical problem. Recent approaches estimate per-pixel
uncertainty from rendering residuals or multiview feature inconsistency to
identify dynamic regions. These methods lack direct motion
evidence, and their accuracy may degrade when reconstruction is inaccurate or
features are ambiguous. In addition, such uncertainty merely downweights
pixels in bundle adjustment (BA), so inaccurate uncertainty estimates can
still impair camera pose estimation and 3D reconstruction. Accordingly, we
present a monocular RGB SLAM system for dynamic scenes. Specifically, a
motion uncertainty estimation method is proposed to derive
uncertainty from the residual between optical flow and
camera motion induced flow, directly reflecting object motion. The estimation
varies with frame alignment accuracy, reducing false detections and yielding accurate 
pixel-level masks of dynamic regions. To make BA more robust, we estimate camera poses from high confidence pixels. These poses are then used to reproject
dynamic pixels during joint refinement of poses and depth, reducing
the influence of object motion on both pose and depth updates. Moreover, we
provide a quantitative evaluation of dynamic masks and analyze how
these masks affect the downstream tracking and reconstruction of dynamic
objects. Our approach improves the mean dynamic mask IoU by 35\% over a
state-of-the-art baseline on DAVIS, and reduces the average ATE by
59\% on Bonn and 19\% on TUM relative to the SLAM backbone used in our system.
The source code will be made publicly available.

\end{abstract}

\section{Introduction}

Monocular simultaneous localization and mapping (SLAM) is a
fundamental problem in robotics, with broad applications in autonomous
navigation \cite{vplslam}, augmented reality
\cite{bbcslam}, and virtual reality
\cite{gevo}.
However, dynamic objects disrupt the geometric consistency across views and may
introduce erroneous correspondences. Such correspondences
degrade camera pose estimation and 3D reconstruction, reducing SLAM accuracy
and robustness in real-world scenes. Dynamic environments therefore remain a
challenge for reliable SLAM. This paper focuses on monocular SLAM
in such environments.

\begin{figure}[t]
    \centering
    \includegraphics[
    width=1.0\columnwidth,
    trim=0 2pt 0 0,
    clip
    ]{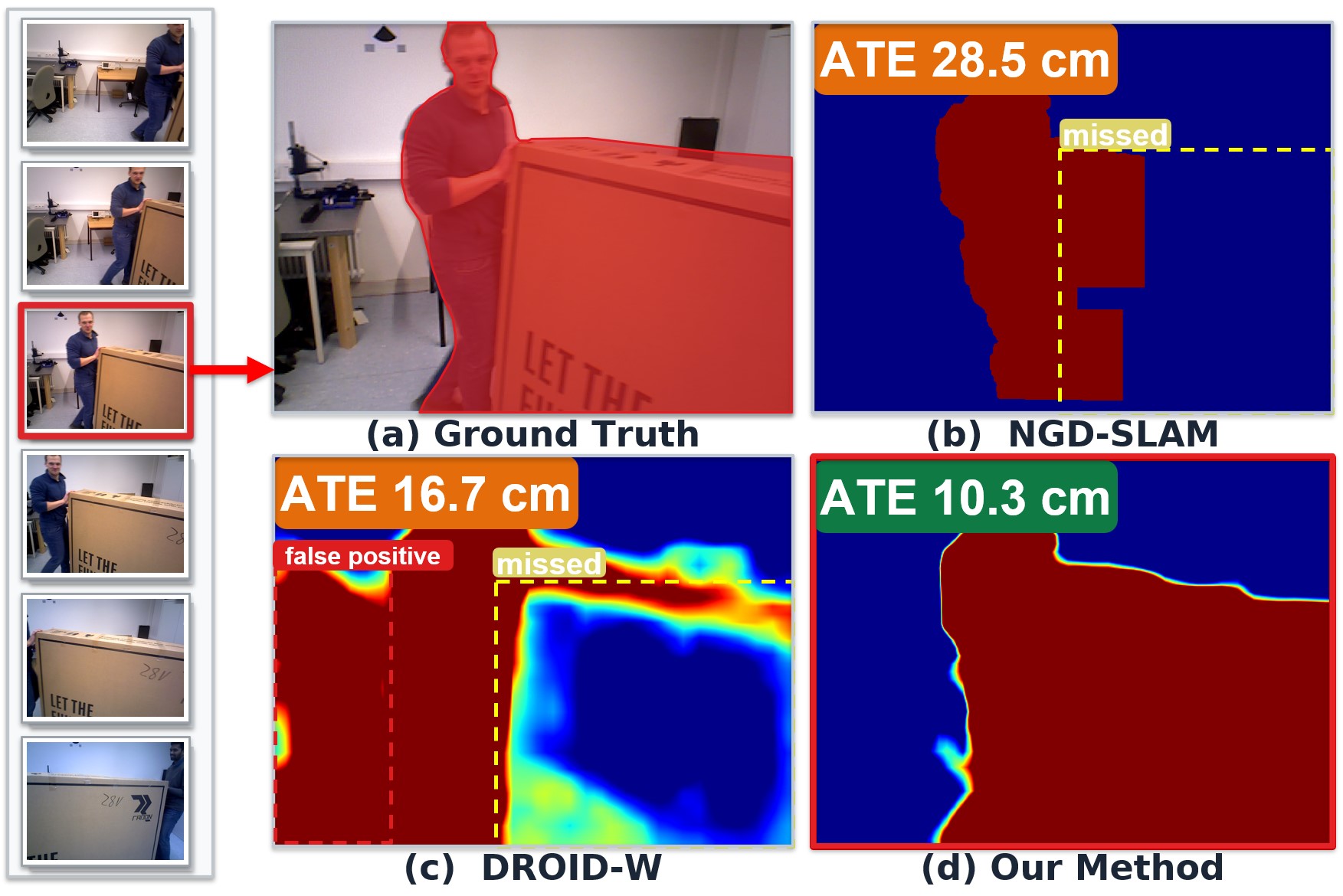}
    \vspace{-20pt}
\caption{\textbf{Motivation.} Dynamic masks and camera localization on
the Bonn sequence. (a) RGB image with the ground-truth dynamic mask. (b) NGD-SLAM~\cite{ngdslam} relies on predefined
object classes and misses part of the dynamic region. (c) DROID-W~\cite{droidw}
infers motion indirectly from feature inconsistency, producing an incomplete
mask and false positives. (d) Our method uses optical flow as direct motion evidence, yielding
more accurate masks and reducing ATE by 63.9\% and 38.3\% compared with NGD-SLAM and DROID-W, respectively.}

\label{fig:motivation}
    \vspace{-2pt}
\end{figure}

Identifying dynamic objects underpins subsequent camera tracking and 3D scene
reconstruction. Moving object segmentation (MOS) models
\cite{seganymo,geomotion,gmos} can identify dynamic objects in video, but these models typically have prohibitive inference costs for real-time tasks.
Earlier dynamic SLAM methods instead use lightweight semantic segmentation or
object detection models \cite{dynaslam,dsslam,detectslam}, which infer motion from
semantic categories rather than actual motion. As a result, they fail to
recognize moving objects from unseen classes. For known classes, they may
mistake static objects for dynamic ones and moving ones for static.
Open-vocabulary segmentation \cite{ovseg} broadens object coverage but
leaves this semantic basis unchanged. Fig.~\ref{fig:motivation} illustrates
the potential errors of the detection-based method NGD-SLAM \cite{ngdslam}.
                                                                                                                           
Recent works \cite{wildgsslam,upslam} estimate per-pixel uncertainty for
dynamic regions, independent of object categories. This uncertainty is
predicted from DINO \cite{dinov2} features with a shallow
multilayer perceptron (MLP), supervised online by the discrepancy between
input images and current renderings. The estimates degrade when the
map is inaccurate. DROID-W \cite{droidw} avoids this dependence 
on reconstruction accuracy by using multiview feature inconsistency.  
However, the feature inconsistency provides indirect evidence of motion
and may be unreliable in semantically complex or weakly textured regions,
as shown in the second row of Fig.~\ref{fig:motivation}.

Optical flow captures interframe motion directly, and several
SLAM systems use it to detect dynamic regions
\cite{flowfusion,dsslam,rodynslam}. In these systems, however, the flow comes
from a dedicated estimator, adding computational overhead and increasing
per-frame processing time \cite{dynagslam}.
In addition, the identification scheme for
marking a pixel as dynamic is fixed in advance
\cite{flowfusion,dynamon,d3flowslam}. Consequently, pose and depth errors
may cause static pixels to be misclassified as dynamic. 

Apart from dynamic object identification, how the estimated uncertainty is used in
optimization also affects system robustness. Existing uncertainty-aware
systems \cite{wildgsslam,droidw} apply uncertainty estimates in
bundle adjustment (BA) as per-pixel weights, and refine poses and depths
jointly under that weighting. However, motion residuals from dynamic pixels
that are not fully suppressed still impact both depth estimates and keyframe
poses. All pixels are used in BA, so inaccurate uncertainty at any pixel can affect
the estimation.

Motivated by the above observations, we present a monocular RGB SLAM system for dynamic scenes built on \textbf{FL}ow-derived uncertainty and \textbf{T}wo-\textbf{S}tage optimization (FLoTS-SLAM). 
Optical flow measures motion directly and is obtained from the existing
correspondence predictions of the system tracker without additional computation.
To identify dynamic objects accurately, we design a motion uncertainty
estimation method based on optical flow residuals.
In addition, uncertainty weighting alone may not fully suppress the influence
of dynamic pixels on pose and depth estimation, and removing these pixels
would reduce the number of constraints for optimization. We therefore perform
optimization in two stages to improve robustness. 
In the first stage, depth is fixed and poses are estimated from high confidence pixels. 
In the second stage, poses and depth are jointly refined, 
using the first-stage poses to predict where dynamic pixels would appear if the scene were static.
Our main contributions are summarized as follows:

\begin{itemize}
    \item We propose a motion uncertainty estimation method
    to improve the accuracy of dynamic object identification. It
    derives uncertainty from the residual between optical flow and
    camera motion induced flow, and adapts to frame alignment accuracy.

    \item We introduce a two-stage optimization method to reduce the influence of
    object motion on pose and depth updates, and improve tracking
    robustness. 

    \item We conduct extensive experiments.
    FLoTS-SLAM attains state-of-the-art camera localization on Bonn
    and TUM, reducing the average ATE of its DROID-SLAM backbone by 59\%
    and 19\%. We also introduce
    Dynamic Mask IoU (DM-IoU) to quantitatively evaluate dynamic identification
    across methods. On DAVIS, our method improves the mean DM-IoU by 35\%
    over a state-of-the-art baseline.

    \item We will make the source code publicly available to facilitate future research.
\end{itemize}

\section{Related Work}

\subsection{Moving Object Segmentation}

Moving object segmentation (MOS) aims to identify dynamic objects in a video and produce masks for each frame.
Early learning-based methods group pixels whose motion is coherent across
frames~\cite{motiongrouping,segmentanythingmoves}, and later work adds complementary cues such as
layered representations, residual-flow modeling, and appearance
features~\cite{oclr,rcf,dpa}. Recent methods instead inherit the generic priors
of foundation models~\cite{seganymo} or ground segmentation in
pretrained 4D geometry~\cite{geomotion,gmos}. All of them, however, run a
dedicated network over a multi-frame window, and their inference costs are
prohibitive for real-time systems.

\subsection{Traditional Dynamic SLAM}

Dynamic objects produce correspondences that rigid camera motion cannot explain,
degrading camera pose estimation and 3D reconstruction in conventional visual
SLAM~\cite{orbslam,droidslam}. Many dynamic SLAM methods therefore detect
dynamic regions and exclude them from optimization, reconstructing the static
scene alone. Such regions are identified from geometric consistency across
frames~\cite{staticfusion,refusion}, from semantic priors on movable
categories~\cite{dsslam,detectslam}, or from a combination of the two.
DynaSLAM~\cite{dynaslam}, for instance, segments movable instances with Mask
R-CNN~\cite{maskrcnn} and verifies them by multiview reprojection consistency.

\subsection{Dynamic NeRF- and 3DGS-based SLAM}

Recent SLAM methods based on neural radiance fields (NeRFs)~\cite{nerf}
and 3D Gaussian Splatting (3DGS)~\cite{gsplat} incorporate dynamic region
identification to handle moving objects. Dynamic extensions such as
PG-SLAM~\cite{pgslam}, NID-SLAM~\cite{nidslam}, and SDD-SLAM~\cite{sddslam}
obtain motion masks from segmentation networks. The masks, however, 
follow object category rather than observed motion, and
cover only the classes the network has seen. 

Another group of methods replaces category priors with a per-pixel uncertainty
map~\cite{nerfonthego}. WildGS-SLAM~\cite{wildgsslam} and UP-SLAM~\cite{upslam}
extend it to SLAM by regressing uncertainty from DINOv2~\cite{dinov2} features,
supervised by the discrepancy between the input image and the current rendering.
The estimate may degrade when the rendering is inaccurate. DROID-SLAM in
the Wild~\cite{droidw} instead derives uncertainty from multiview feature
inconsistency, requiring neither a rendered map nor a learned prior. However, all
of these methods rely on appearance cues, which provide only indirect evidence of
object motion. So our method derives uncertainty from the residual between
optical flow and camera motion induced flow, which measures motion directly.

\subsection{Optimization in Dynamic SLAM}

Optimization in dynamic scenes should account for moving objects.
Early methods such as DynaSLAM~\cite{dynaslam} exclude detected
dynamic regions from tracking and mapping.
DynoSAM~\cite{dynosam} jointly estimates camera poses, scene
structure, and rigid-object motion within a factor graph.
BA-Track~\cite{batrack} learns to isolate camera-induced motion
from observed point trajectories, providing correspondences for BA.
Recent methods~\cite{wildgsslam,droidw} use per-pixel uncertainty to weight residuals
during joint pose and depth refinement.
However, inaccurate weights may insufficiently suppress
motion-induced residuals, compromising estimation accuracy.
We therefore introduce a two-stage optimization method
for more robust and accurate pose and depth estimation.

\section{Method}
\begin{figure*}[t]
  \centering
  \includegraphics[width=1.0\textwidth]{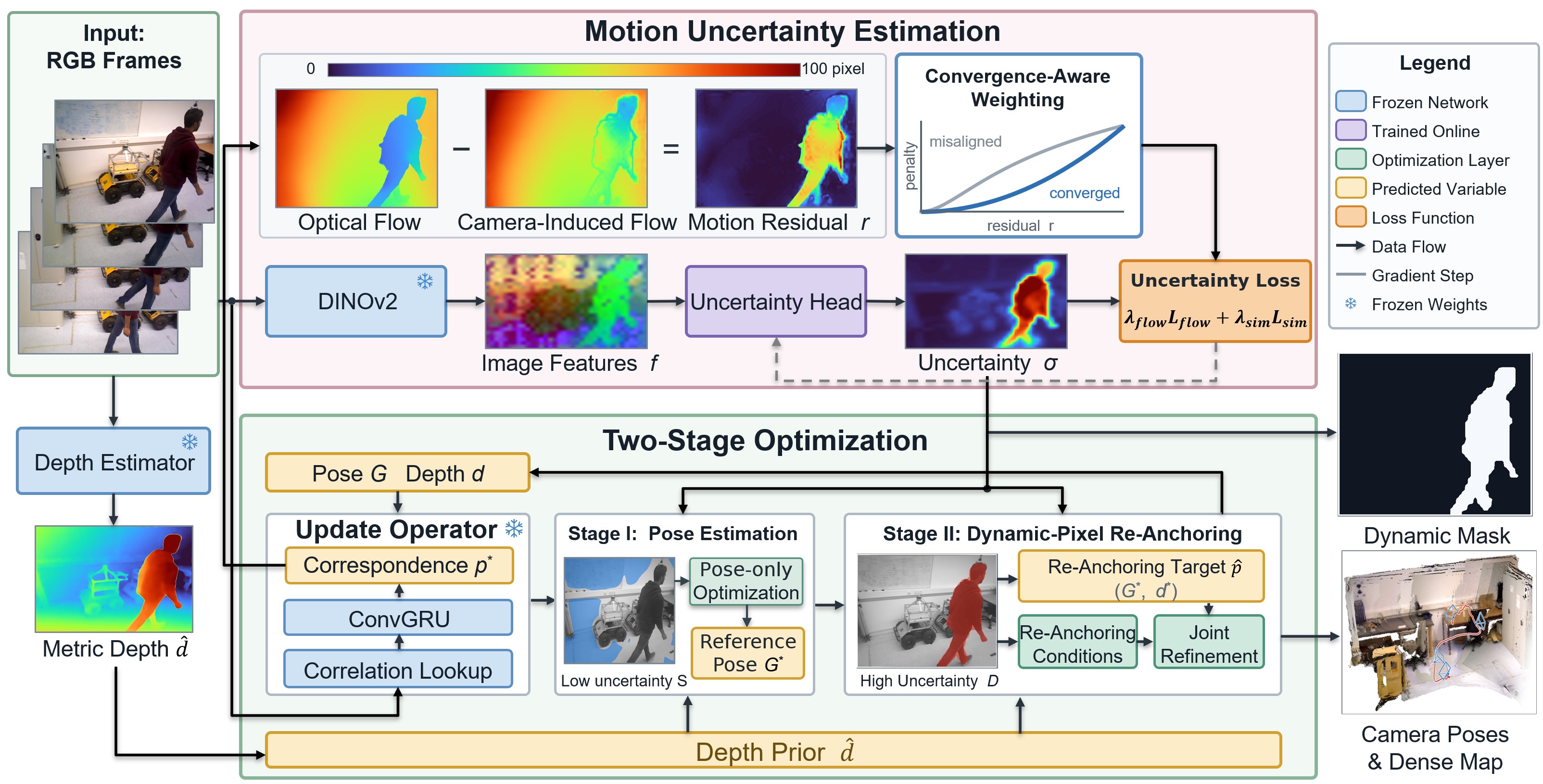}
  \vspace{-0.5cm}
  \caption{\textbf{System Overview.}
  FLoTS-SLAM takes a sequence of monocular RGB images as input and
  jointly estimates camera poses and scene geometry.
  A pre-trained update operator from DROID-SLAM~\cite{droidslam} predicts the
  dense correspondence target $\mathbf{p}^{*}$ and its confidence $\mathbf{W}$.
  DINOv2~\cite{dinov2} extracts the features $\mathbf{f}$, and
  Metric3D~\cite{metric3d} predicts the depth prior $\hat{\mathbf{d}}$.
  All three networks remain frozen.
  In the uncertainty branch (Sec.~\ref{sec:flowunc}), the per-pixel uncertainty
  $\boldsymbol{\sigma}$ is derived from the residual between the optical flow
  and the camera motion induced flow. The estimated uncertainty varies with the
  frame alignment accuracy.
  In the tracking branch (Sec.~\ref{sec:twostage}), the pose is first estimated
  from low uncertainty pixels $\mathcal{S}$ with the depths fixed. The high
  uncertainty pixels $\mathcal{D}$ are then reprojected using the first-stage
  pose before joint refinement. The two branches alternate within each
  Gauss--Newton iteration.}
  \label{fig:overview}
\end{figure*}

\subsection{System Overview}

Fig.~\ref{fig:overview} illustrates the overall architecture of FLoTS-SLAM.
Given a monocular RGB sequence, our system estimates camera poses and
scene depth for robust tracking and dense reconstruction in dynamic
environments (Sec.~\ref{sec:prelim}).
To identify moving objects from motion rather than from appearance,
we derive motion uncertainty from the residual between optical flow
and camera motion induced flow (Sec.~\ref{sec:flowunc}).
The estimation varies with frame alignment accuracy to reduce false
motion detections caused by misalignment.
For robust tracking, we perform optimization in two stages, both using the
estimated uncertainty (Sec.~\ref{sec:twostage}).
The first stage estimates poses from low uncertainty pixels with
depths fixed.
The second stage jointly refines poses and depths, replacing the
correspondence targets of high uncertainty pixels with reprojections
computed using the first-stage poses.
This reduces the influence of object motion on both pose and depth
updates.

\subsection{Dense Tracking with Per-Pixel Uncertainty}
\label{sec:prelim}

\noindent\textbf{Differentiable Dense Bundle Adjustment.}
Given an RGB sequence $\{\mathbf{I}_i\}_{i=0}^{N}$ with each frame of resolution $H\times W$, the tracker maintains
a frame graph $(\mathcal{V},\mathcal{E})$, where
$\mathcal{V}\subseteq\{0,\ldots,N\}$ denotes the set of keyframe indices and each edge
$(i,j)\in\mathcal{E}$ connects two co-visible keyframes.
For each keyframe $i\in\mathcal{V}$, the tracker maintains two state variables, a camera pose
$\mathbf{G}_i\in\mathrm{SE}(3)$ and an inverse depth map
$\mathbf{d}_i\in\mathbb{R}^{(H/8)\times(W/8)}$.
We denote the sets of these variables by
$\mathbf{G}=\{\mathbf{G}_i\}_{i\in\mathcal{V}}$ and
$\mathbf{d}=\{\mathbf{d}_i\}_{i\in\mathcal{V}}$.
For each edge $(i,j)\in\mathcal{E}$, the current estimates induce the correspondence
\begin{equation}
  \mathbf{p}_{ij} \;=\; \Pi_c\!\left(\mathbf{G}_{ij},\,\Pi_c^{-1}(\mathbf{p}_i,\,\mathbf{d}_i)\right),
  \label{eq:reproj}
\end{equation}
where $\Pi_c(\cdot,\cdot)$ is the image projection function,
$\mathbf{G}_{ij}=\mathbf{G}_j\mathbf{G}_i^{-1}$ is the relative transformation from frame $i$
to frame $j$, and $\mathbf{p}_i$ is the pixel grid of frame $i$. For the same edge, a recurrent update operator with fixed pretrained weights
predicts a dense correspondence target $\mathbf{p}^{*}_{ij}$ together with a diagonal confidence matrix
$\mathbf{W}_{ij}$. The target gives the position in frame $j$ where each pixel
of frame $i$ is observed. The states are refined by minimizing
\begin{equation}
  \mathcal{L}_{\mathrm{ba}}(\mathbf{G},\mathbf{d})=\smashoperator{\sum_{(i,j)\in\mathcal{E}}}\|\mathbf{p}^{*}_{ij}-\mathbf{p}_{ij}\|^{2}_{\mathbf{W}_{ij}}+\lambda_d\smashoperator{\sum_{i\in\mathcal{V}}}\|\mathbf{d}_i-\hat{\mathbf{d}}_i\|^{2},
  \label{eq:ba}
\end{equation}
where $\|\cdot\|_{\mathbf{W}}$ denotes the Mahalanobis norm that weights each pixel by its
confidence. The first term is the dense correspondence objective of DROID-SLAM. Monocular
disparity is observable only up to scale. The second term regularizes $\mathbf{d}_i$ toward a
metric prior $\hat{\mathbf{d}}_i$ from Metric3D~\cite{metric3d} with weight $\lambda_d$, after
aligning the prior to the current scale.

Each Gauss--Newton step linearizes~\eqref{eq:ba}, yielding
\begin{equation}
  \begin{bmatrix} \mathbf{B} & \mathbf{E}\\ \mathbf{E}\T & \mathbf{C}\end{bmatrix}
  \begin{bmatrix} \Delta\boldsymbol{\xi}\\ \Delta\mathbf{d}\end{bmatrix}
  =\begin{bmatrix} \mathbf{v}\\ \mathbf{w}\end{bmatrix},
  \label{eq:normal}
\end{equation}
where $\mathbf{B}$ and $\mathbf{C}$ are the pose and depth blocks of the approximate Hessian,
$\mathbf{E}$ is the coupling block, and $\mathbf{v}$ and $\mathbf{w}$ are the
corresponding gradient vectors. $\Delta\boldsymbol{\xi}$ and $\Delta\mathbf{d}$ are the pose and
inverse depth increments.

\noindent\textbf{Uncertainty-Aware Downweighting.}
Equation~\eqref{eq:ba} assumes that every residual comes from errors in the estimated camera poses and inverse depths.
Moving objects violate this assumption because camera motion alone cannot explain their
displacement. Consequently, the solver may incorrectly adjust the camera poses and inverse depths to reduce these motion-induced residuals.
Following~\cite{droidw,wildgsslam}, we attach a per-pixel uncertainty map
$\boldsymbol{\sigma}_i\in\mathbb{R}^{(H/8)\times(W/8)}$ to each keyframe and collect these maps as
$\boldsymbol{\sigma}=\{\boldsymbol{\sigma}_i\}_{i\in\mathcal{V}}$. We then define the
confidence matrix $\boldsymbol{\Omega}_{ij}(\mathbf{u})$ at each pixel as
\begin{equation}
  \boldsymbol{\Omega}_{ij}(\mathbf{u})=g\bigl(\sigma_i(\mathbf{u})\bigr)\,\mathbf{W}_{ij}(\mathbf{u}),
  \label{eq:sigmaw}
\end{equation}
where $\sigma_i(\mathbf{u})$ is the uncertainty at pixel $\mathbf{u}$ of frame $i$ and $g(\cdot)$ is
a monotonically decreasing attenuation function. Pixels with high uncertainty contribute little while remaining in the
frame graph. 

\subsection{Motion Uncertainty Estimation}
\label{sec:flowunc}

\noindent\textbf{Uncertainty Loss Function.}
Consider a pixel $\mathbf{u}$ of source keyframe $i$ on an edge $(i,j)$. The recurrent update
operator gives its correspondence target $\mathbf{p}^{*}_{ij}(\mathbf{u})$ in keyframe $j$,
and~\eqref{eq:reproj} gives the correspondence $\mathbf{p}_{ij}(\mathbf{u})$ induced by camera
motion. The optical flow is $\mathbf{p}^{*}_{ij}(\mathbf{u})-\mathbf{u}$ and the camera motion
induced flow is $\mathbf{p}_{ij}(\mathbf{u})-\mathbf{u}$, and their difference cancels the common
source coordinate,
\begin{equation}
  \mathbf{r}_{ij}\;=\;\mathbf{p}^{*}_{ij}-\mathbf{p}_{ij},
  \label{eq:flowres}
\end{equation}
with all quantities indexed on the pixel grid of keyframe $i$. The residual $\mathbf{r}_{ij}$ represents the component of pixel displacement unexplained by camera motion and scene depth. It
decreases on the static background as the optimization converges, but persists on moving objects.
Writing $r_{ij}(\mathbf{u})=\|\mathbf{r}_{ij}(\mathbf{u})\|$ for
its magnitude, we estimate $\boldsymbol{\sigma}$ using an uncertainty loss motivated by a Gaussian
negative log-likelihood,
\begin{equation}
  \mathcal{L}_{\mathrm{flow}}(\boldsymbol{\sigma})=\sum_{(i,j)\in\mathcal{E}}\sum_{\mathbf{u}}
    \left[\frac{\psi\bigl(r_{ij}(\mathbf{u})\bigr)}{2\,\sigma_i(\mathbf{u})^{2}}+\log\sigma_i(\mathbf{u})\right],
  \label{eq:lflow}
\end{equation}
where $\psi(\cdot)$ adaptively transforms the residual magnitude to control its contribution to the uncertainty loss.
The second term is a regularizer that prevents the trivial minimum at
$\sigma_i(\mathbf{u})\to\infty$ and drives $\sigma_i(\mathbf{u})$ toward the residual level of the
pixel.

Directly using the squared residual $r^{2}$ in the uncertainty loss would be unreliable in early iterations. Errors in the estimated camera poses and inverse depths dominate the residual, and static pixels exhibit large residuals. These large residuals raise the estimated uncertainty of many static pixels. The attenuation in~\eqref{eq:sigmaw} then suppresses correspondences needed for convergence.
Both the residual and the suppression would
persist. We therefore adapt $\psi(\cdot)$ to frame alignment accuracy,
\begin{align}
  \psi(r) &= \beta\,r^{2}+(1-\beta)\log\!\left(1+r^{2}\right),\label{eq:reff}\\
  \beta   &= \operatorname{sigmoid}\bigl(\tau_{r}-\bar{r}\bigr),\label{eq:beta}
\end{align}
where $\bar{r}$ is the median of $r_{ij}(\mathbf{u})$ over all pixels in the current optimization
window, and $\tau_{r}$ is the alignment level of the transition. While the alignment is poor,
$\beta$ approaches zero and the logarithm compresses the residual, so a misaligned pixel is not
classified as dynamic. As the alignment improves, $\beta$ approaches one and $\psi(r)$ approaches
$r^{2}$. When static pixels account for a substantial proportion of the residual samples, the median
can indicate background alignment.

The cost $\mathcal{L}_{\mathrm{flow}}$ treats each pixel independently and lacks
object-level coherence. We complement it with the DINOv2-based feature-consistency cost
$\mathcal{L}_{\mathrm{sim}}$~\cite{droidw}, replacing its similarity threshold with weighting so that the uncertainty varies continuously with the feature similarity. The total
uncertainty cost is
\begin{equation}
  \mathcal{L}_{\mathrm{unc}}=\lambda_{\mathrm{sim}}\,\mathcal{L}_{\mathrm{sim}}+\lambda_{\mathrm{flow}}\,\mathcal{L}_{\mathrm{flow}},
  \label{eq:lunc}
\end{equation}
where $\lambda_{\mathrm{sim}}$ weights the feature-consistency term, while
$\lambda_{\mathrm{flow}}$ weights the flow-residual term.

\noindent\textbf{Regularization and Optimization.}
Directly optimizing per-pixel uncertainty can overfit residual noise and produce spatially inconsistent maps.
Following~\cite{droidw}, we instead predict the uncertainty at each pixel from its DINOv2 feature
through a shared head,
$\sigma_i(\mathbf{u})=h_{\boldsymbol{\theta}}\bigl(\mathbf{f}_i(\mathbf{u})\bigr)$. The head
$h_{\boldsymbol{\theta}}(\cdot)$ is an affine map followed by an offset softplus, which keeps
$\boldsymbol{\sigma}$ bounded away from zero. The parameters $\boldsymbol{\theta}$ are shared across
the sequence, encouraging consistent uncertainty predictions for pixels with similar DINOv2 features.

To keep the uncertainty consistent with the latest camera poses and inverse depths, we alternate between
updating the geometric variables and updating the uncertainty prediction head at every iteration~\cite{droidw}.
Each Gauss--Newton step updates the camera poses $\mathbf{G}$ and inverse depths $\mathbf{d}$
using the objective $\mathcal{L}_{\mathrm{ba}}$, while keeping the uncertainty maps $\boldsymbol{\sigma}$ fixed.
A gradient step then updates the head parameters $\boldsymbol{\theta}$ using the objective
$\mathcal{L}_{\mathrm{unc}}$, while keeping $\mathbf{G}$ and $\mathbf{d}$ fixed.

\subsection{Two-Stage Optimization with Dynamic-Pixel Re-Anchor}
\label{sec:twostage}

Uncertainty maps are not always exact, so the uncertainty attenuation may not fully
suppress the residuals of dynamic pixels. Because object motion induces coherent pixel displacements,
these residuals are systematic offsets rather than zero-mean noise. Least squares does not suppress
such offsets. Under the reprojection geometry, these residuals decompose into an epipolar component, absorbed by the
inverse depth of the pixel, and a perpendicular component, which biases the keyframe pose shared by
all pixels. The depth error persists and can impact pose estimates
on edges with a different baseline, through the coupling in $\mathbf{E}$. So a high
uncertainty pixel should not update the shared pose and the per-pixel depths simultaneously. We thus optimize in two stages: pose estimation and joint refinement.

\noindent\textbf{Stage I: Pose Estimation from Low Uncertainty Pixels.}
Let $\mathcal{S}_{i}$ denote the set of pixels whose uncertainty falls in the lowest quantile of
$\boldsymbol{\sigma}_i$, which are the pixels most likely to be static. Stage I estimates the camera poses
from $\mathcal{S}_{i}$ alone, with all inverse depths held fixed,
\begin{equation}
  \mathbf{G}^{*}=\operatorname*{arg\,min}_{\mathbf{G}}\ \sum_{(i,j)\in\mathcal{E}}\ \sum_{\mathbf{u}\,\in\,\mathcal{S}_{i}}
    \bigl\|\mathbf{p}^{*}_{ij}(\mathbf{u})-\mathbf{p}_{ij}(\mathbf{u})\bigr\|^{2}_{\mathbf{W}_{ij}(\mathbf{u})},
  \label{eq:stage1}
\end{equation}
where $\mathbf{W}_{ij}(\mathbf{u})$ is the confidence predicted by the update operator, used here
without uncertainty attenuation. In Stage I, uncertainty is used to select $\mathcal{S}_{i}$,
while the selected residuals retain their original confidence weights without further uncertainty attenuation.
We restrict $\mathcal{S}_{i}$ to pixels with the lowest uncertainty to reduce the risk of including dynamic pixels.
Holding the depths fixed sets $\Delta\mathbf{d}\equiv0$ and reduces the normal equations to
$\mathbf{B}\Delta\boldsymbol{\xi}=\mathbf{v}$, so the selected pixels constrain only the pose
rather than both the pose and their own geometry. Stage I imposes a less stringent requirement on
$\boldsymbol{\sigma}$ than uncertainty attenuation. It need not be accurate everywhere, only
accurate enough to exclude dynamic pixels from the lowest quantile. The result $\mathbf{G}^{*}$ is
not the final pose, but a reference pose obtained without any contribution from the moving parts of
the scene.

\noindent\textbf{Stage II: Dynamic-Pixel Re-Anchoring.}
Stage II refines the poses and the inverse depths jointly over all pixels, using the uncertainty
weights $\boldsymbol{\Omega}_{ij}$ and with $\mathbf{G}^{*}$ as the initial pose. Down-weighting leaves
the correspondence target of a high uncertainty pixel unchanged, so the solver can still adjust the
inverse depth of that pixel toward a target that camera motion cannot explain. We therefore replace
the target itself, and refer to this replacement as dynamic-pixel
re-anchoring. Let $\mathcal{D}_{i}$ denote the set of pixels whose uncertainty falls in the highest
quantile of $\boldsymbol{\sigma}_i$. For these pixels, the correspondence target is replaced by the
prediction of the reference pose,
\begin{align}
  \hat{\mathbf{p}}_{ij} &= \Pi_c\!\left(\mathbf{G}^{*}_{ij},\,\Pi_c^{-1}(\mathbf{p}_i,\,\mathbf{d}^{*}_i)\right),
    \label{eq:lockedpred}\\
  \tilde{\mathbf{p}}^{*}_{ij}(\mathbf{u}) &=
    \begin{cases}
      \hat{\mathbf{p}}_{ij}(\mathbf{u}), & \mathbf{u}\in\mathcal{D}_{i},\\[2pt]
      \mathbf{p}^{*}_{ij}(\mathbf{u}),   & \text{otherwise},
    \end{cases}
    \label{eq:reanchor}
\end{align}
where $\hat{\mathbf{p}}_{ij}$ is the correspondence predicted by the reference pose
$\mathbf{G}^{*}$ and the depths $\mathbf{d}^{*}$ that Stage I held fixed.
It is computed once and then remains fixed throughout Stage II.
With the modified targets and uncertainty weights,
Stage II jointly optimizes the poses and inverse depths,
\begin{equation}
  \mathcal{L}_{\mathrm{st2}}(\mathbf{G},\mathbf{d})=\smashoperator{\sum_{(i,j)\in\mathcal{E}}}\|\tilde{\mathbf{p}}^{*}_{ij}-\mathbf{p}_{ij}\|^{2}_{\boldsymbol{\Omega}_{ij}}+\lambda_d\smashoperator{\sum_{i\in\mathcal{V}}}\|\mathbf{d}_i-\hat{\mathbf{d}}_i\|^{2}.
  \label{eq:stage2}
\end{equation}
Intuitively, a re-anchored pixel is matched not to where it was observed,
but to where it would have appeared if the scene were static.

The re-anchored residual $\hat{\mathbf{p}}_{ij}-\mathbf{p}_{ij}$ vanishes at the reference state, where $\mathbf{G}=\mathbf{G}^{*}$ and $\mathbf{d}=\mathbf{d}^{*}$.
Consequently, this residual does not affect the gradient vectors $\mathbf{v}$ and $\mathbf{w}$ when the objective is linearized at this state.
This removes the gradient that object motion would otherwise contribute to the shared pose and to
the inverse depth of the pixel. The re-anchored residual term still contributes to the approximate Hessian blocks
$\mathbf{B}$, $\mathbf{E}$, and $\mathbf{C}$, retaining a local quadratic constraint around the reference state during joint refinement.

\noindent\textbf{Conditions for Re-Anchoring.}
Re-anchoring inherits all errors in $\mathbf{G}^{*}$, so two conditions decide whether it is
applied. The first compares the departure of the correspondence targets from the reference-pose
prediction on $\mathcal{D}_{i}$ and on $\mathcal{S}_{i}$. We apply re-anchoring only when the discrepancy is substantially
greater for $\mathcal{D}_{i}$ than for $\mathcal{S}_{i}$. A weak contrast indicates either no distinct object
motion or an insufficiently accurate $\mathbf{G}^{*}$. The second concerns online tracking, where
corrections with consistent directions accumulate along the trajectory. During a warm-up phase we
compare the camera centers obtained with and without re-anchoring. Once the corrections are
directionally consistent, we re-anchor only in the final global adjustment. When re-anchoring is not applied, 
Stage II keeps the original targets $\mathbf{p}^{*}_{ij}$.

\begin{figure*}[!t]
  \centering
  \includegraphics[width=0.95\textwidth]{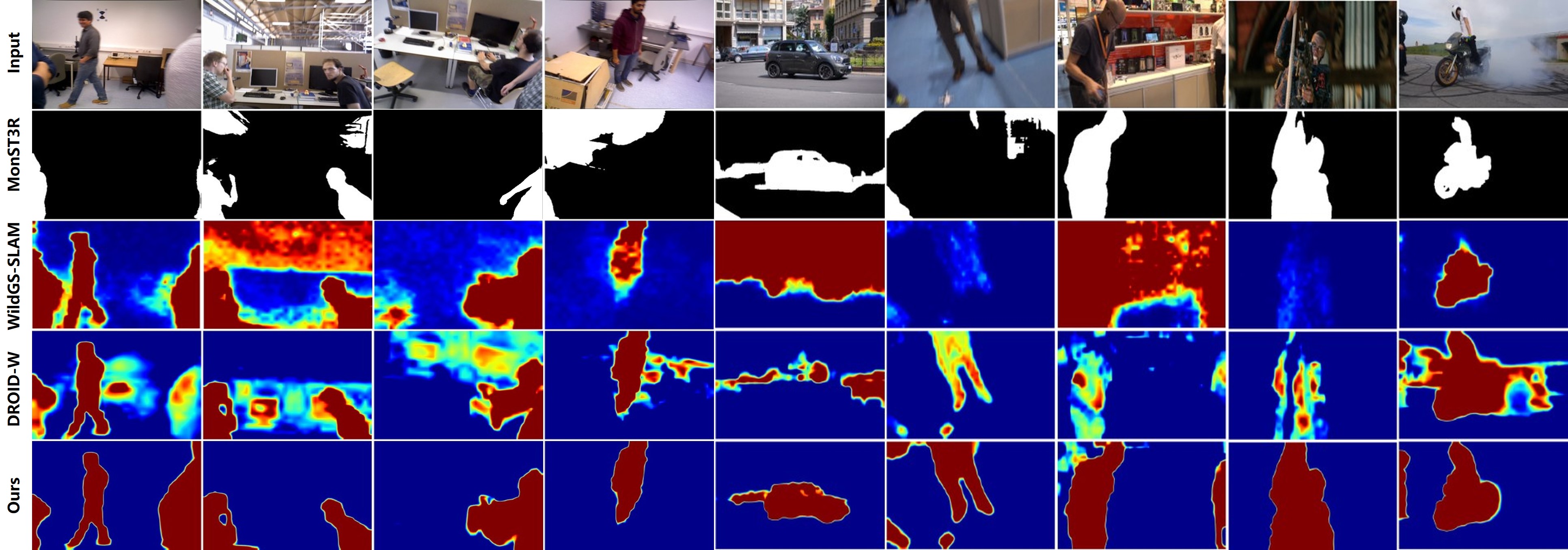}
  \vspace{-3pt}
  \caption{\textbf{Qualitative comparison of dynamic mask predictions on the Bonn, TUM-RGBD, and DAVIS datasets.}
  The rows show, from top to bottom, the input RGB frame and the predictions of
  MonST3R~\cite{monst3r}, WildGS-SLAM~\cite{wildgsslam},
  DROID-W~\cite{droidw}, and FLoTS-SLAM.
  The left four columns are indoor scenes from Bonn and TUM-RGBD, and the right five columns
  are DAVIS sequences.
  Our method produces spatially coherent masks with sharper boundaries while suppressing
  false responses.}
  \label{fig:dynmask}
\end{figure*}

\section{Experiments}

\textbf{Datasets.}  We evaluate on three public benchmarks. The Bonn RGB-D
Dynamic dataset~\cite{refusion} and the dynamic sequences of TUM RGB-D~\cite{tumrgbd} assess camera
localization robustness under moving foreground interference. DAVIS~\cite{davis}
provides pixel-level annotations of moving objects for evaluating the quality of the predicted dynamic masks.

\textbf{Baselines.} We compare with baselines from four categories:
(a)~semantic-/geometry-based dynamic SLAM: DynaSLAM\,(N+G)~\cite{dynaslam};
(b)~general-purpose deep SLAM: DROID-SLAM~\cite{droidslam};
(c)~dynamic Gaussian-splatting SLAM: DG-SLAM~\cite{dgslam}, UP-SLAM~\cite{upslam}, ADD-SLAM~\cite{addslam},
Flow4DGS-SLAM~\cite{flow4dgsslam}, and WildGS-SLAM~\cite{wildgsslam};
(d)~robust dynamic-scene methods: DynaMoN\,(MS\&SS)~\cite{dynamon} and DROID-W~\cite{droidw}.
We also include MonST3R~\cite{monst3r} as a feed-forward baseline.

\textbf{Metrics.}
We evaluate the accuracy of dynamic object identification and camera localization,
as well as processing speed. For localization, we report the Absolute Trajectory Error (ATE RMSE)~\cite{tumrgbd} after
$\mathrm{Sim}(3)$ alignment.
For efficiency, we report frames per second (FPS).
We introduce Dynamic Mask IoU (DM-IoU) as the spatial agreement between
predicted and ground-truth dynamic masks,
\begin{equation}
\mathrm{DM\text{-}IoU}=\frac{|\widehat{\mathcal{M}}_{\mathrm{dyn}}\cap\mathcal{M}_{\mathrm{dyn}}^{\mathrm{gt}}|}{|\widehat{\mathcal{M}}_{\mathrm{dyn}}\cup\mathcal{M}_{\mathrm{dyn}}^{\mathrm{gt}}|},
\end{equation}
where $\widehat{\mathcal{M}}_{\mathrm{dyn}}$ and
$\mathcal{M}_{\mathrm{dyn}}^{\mathrm{gt}}$ denote predicted and ground-truth
masks. A high DM-IoU requires both complete coverage and low background
contamination.

\begin{table}[!t]
\centering
\makeatletter
\long\def\@makecaption#1#2{%
  {\centering
   \footnotesize
   #1\par
   {\scshape #2\par}%
  }%
  \vskip 6pt
}
\makeatother
\caption{Quantitative comparison of DM-IoU on DAVIS scene types.}
\label{tab:davis_scene_type}
\footnotesize
\renewcommand{\arraystretch}{1.14}
\setlength{\tabcolsep}{2.0pt}
\begin{tabular}{llcccc}
\toprule
\textbf{Type} & \textbf{Seq.} & \textbf{MonST3R} & \textbf{\makecell{WildGS\\[-2pt]SLAM}} & \textbf{\makecell{DROID\\[-2pt]W}} & \textbf{Ours} \\
\midrule
\multirow{4}{*}{\makecell[l]{Ani-\\mals}}
& camel      & 0.23 & 0.00 & 0.66 & \cellcolor{bestblue}\textbf{0.74} \\
& cows       & 0.04 & 0.27 & 0.75 & \cellcolor{bestblue}\textbf{0.84} \\
& koala      & 0.13 & 0.36 & \cellcolor{bestblue}\textbf{0.76} & 0.59 \\
\cdashline{2-6}
& mean       & 0.13 & 0.21 & \cellcolor{bestblue}\textbf{0.73} & 0.72 \\
\midrule
\multirow{4}{*}{\makecell[l]{Pedes-\\trians}}
& drone      & 0.47 & 0.32 & 0.36 & \cellcolor{bestblue}\textbf{0.75} \\
& lucia      & \cellcolor{bestblue}\textbf{0.78} & 0.19 & 0.64 & 0.69 \\
& color-run  & 0.47 & 0.31 & 0.55 & \cellcolor{bestblue}\textbf{0.68} \\
\cdashline{2-6}
& mean       & 0.57 & 0.27 & 0.51 & \cellcolor{bestblue}\textbf{0.71} \\
\midrule
\multirow{4}{*}{\makecell[l]{Human\\Motion}}
& dance-twirl & 0.60 & 0.48 & \cellcolor{bestblue}\textbf{0.70} & \cellcolor{bestblue}\textbf{0.70} \\
& lindy-hop  & 0.34 & 0.47 & 0.66 & \cellcolor{bestblue}\textbf{0.74} \\
& shooting   & 0.67 & 0.42 & 0.11 & \cellcolor{bestblue}\textbf{0.72} \\
\cdashline{2-6}
& mean       & 0.54 & 0.46 & 0.49 & \cellcolor{bestblue}\textbf{0.72} \\
\midrule
\multirow{4}{*}{\makecell[l]{Wheel.\\Board}}
& longboard  & 0.38 & 0.47 & 0.75 & \cellcolor{bestblue}\textbf{0.84} \\
& mbike-trick & 0.60 & \cellcolor{bestblue}\textbf{0.61} & 0.40 & 0.59 \\
& horsejump-h & \cellcolor{bestblue}\textbf{0.76} & 0.71 & 0.55 & 0.70 \\
\cdashline{2-6}
& mean       & 0.58 & 0.60 & 0.57 & \cellcolor{bestblue}\textbf{0.71} \\
\midrule
\multirow{4}{*}{Vehicles}
& car-round  & 0.70 & 0.21 & 0.10 & \cellcolor{bestblue}\textbf{0.84} \\
& night-race & 0.13 & 0.17 & 0.38 & \cellcolor{bestblue}\textbf{0.78} \\
& car-turn   & 0.61 & 0.38 & \cellcolor{bestblue}\textbf{0.73} & 0.72 \\
\cdashline{2-6}
& mean       & 0.48 & 0.25 & 0.40 & \cellcolor{bestblue}\textbf{0.78} \\
\midrule
\multicolumn{2}{l}{\textbf{Overall Mean}} & 0.46 & 0.36 & 0.54 & \cellcolor{bestblue}\textbf{0.73} \\
\bottomrule
\end{tabular}
\end{table}

\subsection{Evaluation of Dynamic Masks}

We compare dynamic-mask predictions on the Bonn, TUM RGB-D, and DAVIS
datasets in Fig.~\ref{fig:dynmask}. MonST3R sometimes misses parts of moving
objects, while WildGS-SLAM may produce imprecise responses under complex
appearance or rapid motion. DROID-W better identifies the main moving objects
but still responds to visually distinctive static structures and provides
incomplete object coverage. By incorporating an optical-flow motion residual
into uncertainty estimation, our method suppresses these background responses
while producing more complete object regions and sharper boundaries. This
advantage holds not only in the indoor pedestrian scenes of Bonn and TUM but
also generalizes to the objects and motion patterns in DAVIS, including
animals, vehicles, and non-rigid human motion.

DAVIS provides pixel-level ground-truth masks, enabling quantitative evaluation
via DM-IoU. For further analysis, we divide the sequences into five
categories (Animals, Pedestrians, Human Motion, Wheeled and Board Devices,
Vehicles) and report per-category results. Our method
achieves the best overall DM-IoU and leads in most categories, performing
particularly well on human-motion and vehicle scenes. These results confirm the generalization of the
proposed uncertainty estimation across diverse dynamic scenes.

\begin{figure*}[!t]
    \centering
    \includegraphics[width=0.90\textwidth]{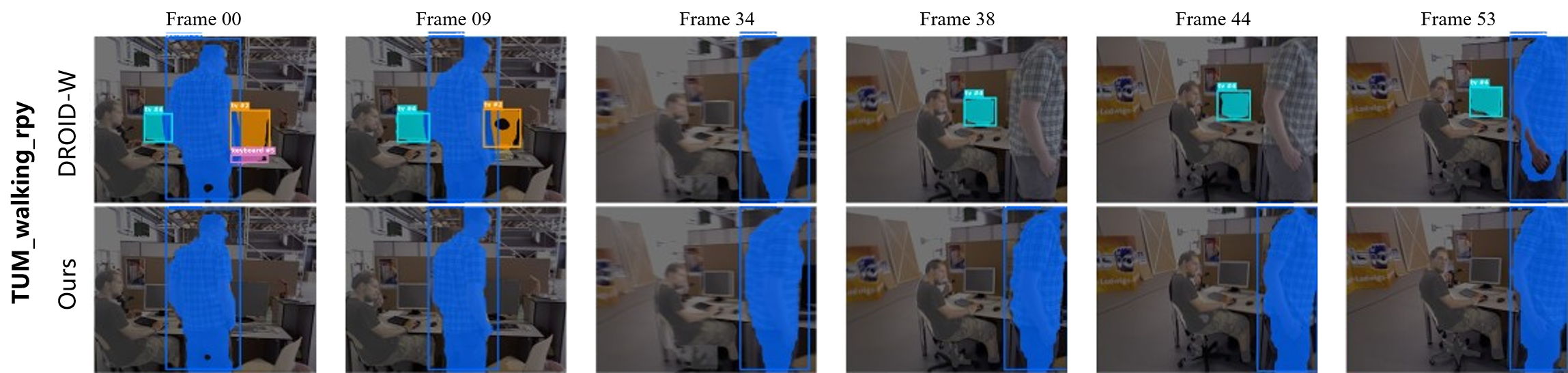}
    \caption{\textbf{Impact of dynamic masks on downstream object tracking.}
    Each method's mask determines which objects a tracker follows as dynamic
    (same color $=$ same track ID). Our masks consistently track moving people,
    while DROID-W responds to static furniture and loses
    continuity on actual movers.}
    \label{fig:tracking_affect}
\end{figure*}
\begin{figure*}[!t]
    \centering
    \includegraphics[width=0.90\textwidth]{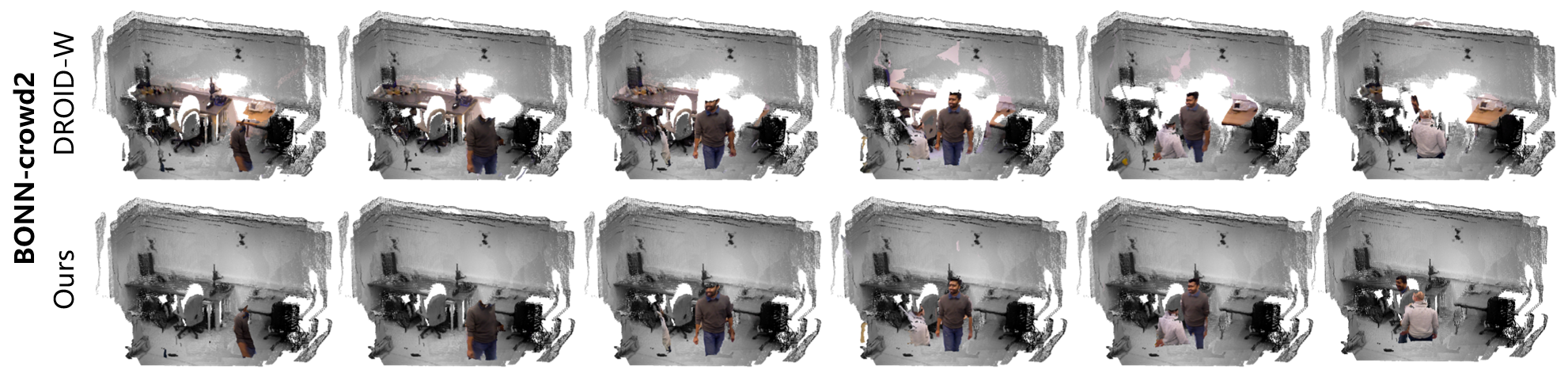}
    \vspace{-3pt}
    \caption{\textbf{Impact of dynamic masks on downstream 3D reconstruction.}
All pixels are back-projected using monocular depth. The static background
is shown in gray, while dynamic regions are colored. Our more complete masks reduce spurious background geometry, 
resulting in more accurate reconstructions.}
    \label{fig:reconstruction_affect}
\end{figure*}

\vspace{10pt}
\begin{table*}[!t]
  \centering
  \makeatletter
  \long\def\@makecaption#1#2{%
    {\centering
     \footnotesize
     #1\par
     {\scshape #2\par}%
    }%
    \vskip 8pt
  }
  \makeatother
  \caption{Camera tracking results (ATE RMSE [cm]) on the Bonn RGB-D Dynamic
  dataset and TUM RGB-D dynamic sequences.}
  \label{tab:bonn}
  \resizebox{\textwidth}{!}{%
  \footnotesize
  \setlength{\tabcolsep}{4.2pt}
  \renewcommand{\arraystretch}{1.14}
  \begin{tabular}{l|ccccccccc|c|ccccccccc|c}
  \hline
  \multicolumn{1}{c}{}
  & \multicolumn{10}{c}{\textbf{Bonn RGB-D Dynamic}}
  & \multicolumn{10}{c}{\textbf{TUM RGB-D}} \\
  \hline
  
  Method
  & Bal. & Bal.2 & Cro. & Cro.2 & Cro.3
  & Per. & Per.2 & Mov. & Mov.2 & Avg.
  & f2/dp & f3/ss & f3/sx & f3/sr & f3/shs
  & f3/ws & f3/wx & f3/wr & f3/whs & Avg. \\
  
  \hline
  \multicolumn{21}{l}{\cellcolor{typegray}\textbf{RGB-D}} \\
  
  DynaSLAM(N+G)
  & 3.0 & 2.9 & 1.6 & 3.1 & 2.7 & 6.1 & 7.8 & 23.2 & 3.9 & 6.04
  & 0.7 & 0.5 & 1.5 & 2.7 & 1.7
  & 0.6 & 1.5 & 3.5 & 2.5 & 1.69 \\
  
  DG-SLAM
  & 3.7 & 4.1 & -- & -- & -- & 4.5 & 6.9 & -- & 3.5 & --
  & 3.2 & -- & 1.0 & -- & --
  & 0.6 & 1.6 & 4.3 & -- & -- \\
  
  UP-SLAM
  & 2.8 & 2.7 & -- & -- & -- & 4.0 & 3.6 & -- & 3.2 & --
  & 1.3 & -- & 0.9 & -- & --
  & 0.7 & 1.6 & -- & 2.6 & -- \\
  
  ADD-SLAM
  & 2.7 & 2.3 & -- & -- & -- & 2.4 & 3.7 & -- & 2.1 & --
  & -- & -- & -- & -- & 1.3
  & 0.5 & 1.4 & -- & 1.6 & -- \\
  
  Flow4DGS
  & 2.6 & 3.4 & 2.0 & 2.8 & 2.1 & 7.2 & 10.8 & 3.9 & 4.3 & 4.33
  & -- & 0.7 & 5.0 & 1.8 & 2.4
  & 0.5 & 3.6 & 2.5 & 4.8 & -- \\
  
  \hline
  \multicolumn{21}{l}{\cellcolor{typegray}\textbf{RGB-only}} \\
  
  MonST3R
  & 7.2 & 6.0 & 6.6 & 6.9 & 4.7 & 9.8 & 16.1 & 3.5 & 6.7 & 7.50
  & 33.9 & 0.8 & 28.3 & 5.1 & 36.8
  & 1.6 & 19.1 & 16.6 & 32.8 & 19.44 \\
  
  DROID-SLAM
  & 7.5 & 4.1 & 5.2 & 6.5 & 6.4 & 4.3 & 5.4 & 2.3 & 4.0 & 5.08
  & \cellcolor{bestblue}\textbf{0.6}
  & \cellcolor{bestblue}\textbf{0.5}
  & 0.9
  & 2.2
  & 1.4
  & 1.2 & 1.6 & 4.0 & 2.2 & 1.62 \\
  
  DynaMoN
  & 2.8 & 2.7 & 3.5 & 2.8 & 2.9 & 14.8
  & \cellcolor{bestblue}\textbf{2.2}
  & \cellcolor{bestblue}\textbf{1.3}
  & 2.7 & 3.96
  & 0.7
  & \cellcolor{bestblue}\textbf{0.5}
  & 0.9
  & 2.4
  & 2.3
  & 0.7 & 1.4
  & 3.9
  & 1.9 & 1.63 \\
  
  WildGS
  & 2.8 & 2.4 & 1.6 & 2.2 & 1.8 & 3.9 & 3.1 & 1.7 & 2.5 & 2.44
  & 1.4
  & \cellcolor{bestblue}\textbf{0.5}
  & \cellcolor{bestblue}\textbf{0.8}
  & 2.4 & 2.0
  & \cellcolor{bestblue}\textbf{0.4}
  & 1.3 & 3.3
  & \cellcolor{bestblue}\textbf{1.6}
  & 1.52 \\
  
  DROID-W
  & 2.6 & 2.5 & 1.3
  & \cellcolor{bestblue}\textbf{1.8}
  & 1.8 & 3.3 & 2.9 & 1.6
  & \cellcolor{bestblue}\textbf{2.3}
  & 2.23
  & 1.1
  & \cellcolor{bestblue}\textbf{0.5}
  & \cellcolor{bestblue}\textbf{0.8}
  & \cellcolor{bestblue}\textbf{2.1} & 1.4 & 0.5
  & \cellcolor{bestblue}\textbf{1.2}
  & 3.1
  & \cellcolor{bestblue}\textbf{1.6}
  & 1.37 \\
  
  \hline
  Ours
  & \cellcolor{bestblue}\textbf{2.4}
  & \cellcolor{bestblue}\textbf{2.1}
  & \cellcolor{bestblue}\textbf{1.2}
  & \cellcolor{bestblue}\textbf{1.8}
  & \cellcolor{bestblue}\textbf{1.6}
  & \cellcolor{bestblue}\textbf{3.1}
  & 2.4
  & 1.6
  & \cellcolor{bestblue}\textbf{2.3}
  & \cellcolor{bestblue}\textbf{2.06}
  & 1.0
  & \cellcolor{bestblue}\textbf{0.5}
  & \cellcolor{bestblue}\textbf{0.8}
  & 2.2
  & \cellcolor{bestblue}\textbf{1.3}
  & 0.5
  & \cellcolor{bestblue}\textbf{1.2}
  & \cellcolor{bestblue}\textbf{2.8}
  & \cellcolor{bestblue}\textbf{1.6}
  & \cellcolor{bestblue}\textbf{1.32} \\
  \hline
  \end{tabular}%
  }
  \label{tab:tum}
  \par\vspace{7pt}
  \end{table*}

\subsection{Impact of Dynamic-Mask Accuracy on Downstream Tasks}

We evaluate how dynamic-mask accuracy affects dynamic-object tracking and
reconstruction. To isolate mask quality, all methods use the same downstream
pipeline. For tracking, Mask R-CNN~\cite{maskrcnn} identifies instances from the binary dynamic masks, and
Hungarian matching associates them across frames.

In tracking, false positives create spurious tracks on static objects, while
incomplete masks break tracks of moving objects. In
Fig.~\ref{fig:tracking_affect}, DROID-W tracks stationary tables and monitors
but loses continuity on people, whereas our masks preserve stable identities
for moving people. In reconstruction, false positives assign static geometry
to the dynamic layer, leaving gaps in the static scene, while missed dynamic
regions yield incomplete object geometry. In
Fig.~\ref{fig:reconstruction_affect}, DROID-W reconstructs walls and desks as
dynamic objects and recovers only partial human geometry, whereas our method
produces compact dynamic-object reconstructions with minimal background
contamination. So dynamic-mask accuracy is critical for
downstream scene understanding.

\subsection{Evaluation of Camera Tracking Performance}

We evaluate camera-tracking accuracy on the Bonn RGB-D Dynamic dataset
and on the dynamic sequences of TUM RGB-D, reporting the ATE RMSE in centimeters
after $\mathrm{Sim}(3)$ alignment in Table~\ref{tab:bonn}. 
Baseline results for individual sequences are taken from
DROID-W~\cite{droidw} where available. The remaining reported baseline
results are obtained from our own runs.

\textbf{Bonn.} Our method attains the lowest average ATE of 2.06\,cm,
ahead of the strongest RGB-only baselines DROID-W (2.23) and WildGS-SLAM
(2.44), and ranks first or tied-first on seven of the nine
sequences. Despite using only monocular RGB input, our method achieves a lower
average ATE than the depth-aided DynaSLAM\,(N+G) (6.04\,cm).
The gains are largest where a prominent moving foreground coexists with
sufficient static structure, so that the two-stage optimization obtains an
accurate reference pose and re-anchoring is effective. On sequences where the
two tests are not satisfied, re-anchoring is disabled and the tracker preserves
the two-stage accuracy without degradation.

\textbf{TUM.} Our method again obtains the lowest average ATE of
1.32\,cm, ahead of DROID-W (1.37) and WildGS-SLAM (1.52). As expected,
the margins are smaller, because one or two walkers in a largely static indoor
scene leave the optimization with abundant static constraints even under
uncertainty down-weighting alone.

\begin{table}[!t]
\centering
\footnotesize
\setlength{\tabcolsep}{4pt}
\setlength{\parskip}{0pt}
\renewcommand{\arraystretch}{1.0}

\makeatletter
\long\def\@makecaption#1#2{%
  {\centering
   \footnotesize
   #1\par
   {\scshape #2\par}%
  }%
  \vskip 10pt
}
\makeatother

\caption{Ablation study on the Bonn RGB-D Dynamic dataset.}
\label{tab:ablation}

\begin{tabular}{lc}
\toprule
Configuration & ATE RMSE [cm] \\
\midrule
a. w/o uncertainty in BA           & 3.62 \\
b. w/o re-anchoring conditions      & 2.37 \\
c. w/o two-stage optimization       & 2.29 \\
d. w/o dynamic-pixel re-anchoring    & 2.23 \\
e. w/ DROID-W uncertainty estimator & 2.16 \\
\midrule
Full (Ours) & \textbf{2.06} \\
\bottomrule
\end{tabular}

\par
\vspace{10pt}
\nointerlineskip

\caption{Runtime comparison in frames per second (FPS).}
\label{tab:runtime}

\begin{tabular}{lcccc}
\toprule
Dataset & DynaMoN & WildGS-SLAM & DROID-W & Ours \\
\midrule
Bonn & 2.4 & 0.3 & \textbf{12.2} & 11.5 \\
TUM  & 2.3 & 0.3 & \textbf{14.4} & 13.0 \\
\bottomrule
\end{tabular}

\vspace{-3pt}
\end{table}

\vspace*{-2pt}
\subsection{Ablation Study}

We ablate each component on nine Bonn sequences
and report the results in Table~\ref{tab:ablation}. Removing any component leads to higher
camera-tracking errors. Disabling uncertainty estimation in BA causes the most
significant performance degradation, demonstrating the importance of suppressing
inconsistent correspondences in dynamic regions. Removing the re-anchoring
conditions, the two-stage optimization, or the dynamic-pixel re-anchoring also consistently
degrades tracking performance, indicating that these components jointly improve
robustness by providing an accurate reference pose and reducing the influence of
object motion on pose optimization. For the experiments \textit{e. w/ DROID-W uncertainty estimator}, 
we replace our uncertainty estimator with that of DROID-W while retaining 
the two-stage optimization and conditional re-anchoring.

\subsection{Runtime Analysis}

All runtime experiments are conducted on a single NVIDIA RTX 4090 GPU
and a 96-core CPU, with the results reported in Table~\ref{tab:runtime}.
The reported tracking time includes monocular depth estimation and
DINOv2 feature extraction. FLoTS-SLAM incurs a modest additional cost
due to the pose-only solve in the first optimization stage and the two
re-anchoring tests. Nevertheless, it maintains a processing rate above 10\,FPS on both
datasets, achieving a favorable balance between robustness and efficiency.

 \vspace*{-2pt}
\section{Conclusion}
\label{sec:conclusion}

In this paper, we propose FLoTS-SLAM, a monocular RGB SLAM system for accurate dynamic object 
identification and robust camera tracking in dynamic environments.
Our method derives motion uncertainty from optical flow residuals and adapts this estimation
to frame alignment accuracy to improve dynamic object identification.
A two-stage optimization method with dynamic-pixel re-anchoring reduces the influence of
object motion on pose and depth updates, thereby improving tracking robustness.
On DAVIS, FLoTS-SLAM improves the mean DM-IoU by approximately 35\% over a
state-of-the-art baseline.
In addition, FLoTS-SLAM reduces the average ATE by approximately 59\% on Bonn and 19\% on TUM
relative to the DROID-SLAM used in our system.

\IEEEtriggeratref{50}

\end{document}